%% file: MainFile.tex
\documentclass[journal,comsoc]{IEEEtran}
\usepackage{epsfig}
\usepackage{graphicx}
\usepackage{amsmath}
\usepackage{amssymb}
\usepackage{multirow}
\usepackage{tabulary}
\usepackage{tabularx}
\usepackage[tight,normalsize,sf,SF]{subfigure}
\usepackage{color}
\usepackage{soul}
\usepackage{enumitem}
\usepackage{hhline}
\usepackage{algorithm}
\usepackage{algpseudocode}
\usepackage{array}
\usepackage{flushend}
\usepackage{multirow}
\usepackage{tabulary}
\usepackage{url}
\usepackage{tabularx}
\usepackage{bm}
\usepackage[tight,normalsize,sf,SF]{subfigure}
\usepackage{color}
\usepackage{csquotes}
\DeclareMathSymbol{\Lambda}{\mathalpha}{operators}{3}
\DeclareMathSymbol{\Pi}{\mathalpha}{operators}{5}
\newcolumntype{P}[1]{>{\centering\arraybackslash}p{#1}}
\graphicspath{{./figs/}}
\DeclareGraphicsExtensions{.eps,.ps,.pdf,.png,.tiff}
\renewcommand{\baselinestretch}{0.98}

\DeclareMathOperator*{\argmax}{arg\,max}

\usepackage[T1]{fontenc}
\begin{document}

\title{Diversity-aware Multi-Video Summarization}

\author{Rameswar~Panda,~Niluthpol~Chowdhury~Mithun,
        and~Amit~K.~Roy-Chowdhury,~\IEEEmembership{Senior~Member,~IEEE}
\thanks{$\bullet$ Rameswar Panda, Niluthpol Chowdhury Mithun and Amit~K.~Roy-Chowdhury are with the Department of Electrical and Computer Engineering, University of California, Riverside, CA 925024, USA. 
E-mails: (rpand002@ucr.edu, nmithun@ece.ucr.edu, amitrc@ece.ucr.edu)}
}

\markboth{IEEE Transactions on Image Processing,~Vol.~xx, No.~xx, August~20xx}%
{Shell \MakeLowercase{\textit{et al.}}: Bare Demo of IEEEtran.cls for IEEE Communications Society Journals}

\maketitle

\begin{abstract}
Most video summarization approaches have focused on extracting a summary from a single video; we propose an unsupervised framework for summarizing a collection of videos. We observe that each video in the collection may contain some information that other videos do not have, and thus exploring the underlying complementarity could be beneficial in creating a diverse informative summary. We develop a novel diversity-aware sparse optimization method for multi-video summarization by exploring the complementarity within the videos. Our approach extracts a multi-video summary which is both interesting and representative in describing the whole video collection. To efficiently solve our optimization problem, we develop an alternating minimization algorithm that minimizes the overall objective function with respect to one video at a time while fixing the other videos. Moreover, we introduce a new benchmark dataset, Tour20, that contains 140 videos with multiple human created summaries, which were acquired in a controlled experiment. Finally, by extensive experiments on the new Tour20 dataset and several other multi-view datasets, we show that the proposed approach clearly outperforms the state-of-the-art methods on the two problems---topic-oriented video summarization and multi-view video summarization in a camera network.      

\end{abstract}


\IEEEpeerreviewmaketitle

\section{Introduction}
\label{sec:Introduction}
\input{Introduction}

\vspace{-2mm}
\section{Related Work}
\label{sec:Related Work}
\input{RelatedWork}
\vspace{-2mm}
\section{Diversity-aware Multi-Video Summarization}
\label{sec:Multi-Video Summarization}
\input{Methodology}

\section{Discussions}
\label{sec:Discussions}
\input{Discussions}

\section{Experiments}
\label{sec:Experiments}
\input{Experiments}

\section{Conclusions and Future Works}
\label{sec:Conclusions}
\input{Conclusions}

\section*{Acknowledgment}

This work was partially supported by NSF grant IIS-1316934. We would like to thank Andrew Kwon and Daniel Handojo, two current UCR undergraduate students, for helping in the annotation of the Tour20 dataset. 

{
	\bibliographystyle{ieee}
	\bibliography{egbib,egbib_2}
}

\end{document}

%% file: Introduction.tex
\IEEEPARstart{W}{ith} the recent explosion of big video data, it is becoming increasingly important to automatically extract a brief yet informative summary of these videos in order to enable a more efficient and engaging viewing experience. 
As a result, \textit{video summarization}, that automates this process, has attracted intense attention in the recent years.

Although video summarization has been extensively studied during the past few years, many previous methods mainly focused on summarizing a \textit{single video} by developing a variety of selection criteria (e.g., representativeness~\cite{Ehsan2012,Eric2014,Scalable2012}, interestingness~\cite{Att2005,LucVanGool2014}) to prioritize frames/segments for the output summary. Another important problem and rarely addressed in this context is to find an informative summary from \textit{multiple} videos. Similar to single video summarization problem, the \textit{multi-video summarization} approach seeks to take a set of related videos and extracts key frames/video skims that presents the most important portions of the input videos within a short duration. Application areas include any scenarios where the user is confronted with watching or browsing a set of related videos, like videos given by a search~\cite{li2010multi,wang2009multi,zhang2015effective} or videos captured with multiple video sensors in a camera network~\cite{MultiviewTMM2010,SanjaySir2015,panda2016embedded,panda2016video}. Given that browsing through all the videos is a very time consuming task, we want to explore whether we can automatically create a video summary that can describe the whole video collection within a short duration.

Multi-video summarization is related to the general problem of single-video summarization with two important distinctions. First, these videos are topically related and hence inter-video statistical dependencies need to be properly exploited for obtaining an informative and diverse summary.
Second, different environmental factors like difference in illumination, pose and synchronization
issues across the multiple topic-related videos also pose a challenge in summarizing such videos.
Thus, direct use of methods that attempt to extract summary from single videos may not produce an optimal set of representatives while summarizing multiple topic-related videos.

To address the challenges encountered in a multi-video setting, we propose a Diversity-aware Multi-Video Summarization (DiMS) approach to generate an informative summary by exploring the complementarity between a set of videos.
We observe that each video in the set may contain some information that other videos do not have, and thus exploring the underlying complementarity is of great importance for the success of multi-video summarization.
We achieve this by developing a novel sparse optimization that jointly summarizes a set of videos to find a single summary that can optimally describe the video collection. Our summarization approach consider two aspects. One, it considers \enquote{interestingness} prior in the sparse representative selection to extract summary that is both interesting and representative of the input video. In particular, segments with high interestingness score are more likely to be selected as key video segments compared to the segments with low interestingness score. Second, we introduce a diversity regularizer in the optimization framework to explore the complementarity within multiple videos in extracting a high quality multi-video summary. We finally develop an efficient alternating minimization algorithm to solve our optimization problem. 
Furthermore, rather than manually evaluating the produced summaries, we introduce a new benchmark dataset with multiple ground truth summaries for each video as well as for the video collection. This data allows to asses the performance of any single-video or multi-video summarization algorithm in a fast and repeatable manner. 

\vspace{1mm}
\underline{\textit {Contributions:}} We address an important, and practical problem in this paper---how to extract an informative yet diverse video summary from a collection of videos. Towards solving this problem, we make the following contributions. (1) we propose an unsupervised approach for multi-video summarization by exploring the complementarity within a set of videos; (2) we develop a novel diversity-aware sparse optimization method that can be efficiently solved by an alternating minimization algorithm; (3) we introduce a new dataset, Tour20, along with clear ground truth summaries to evaluate summarization algorithms in a fast and repeatable manner. To the best of our knowledge, this is the biggest dataset for summarization available. (4) we show the effectiveness of our approach in two tasks---topic-oriented video summarization and multi-view video summarization in a camera network. With extensive experiments on both Tour20 and several standard multi-view datasets, we show the superiority of our approach over competing methods for both of the tasks.

%% file: RelatedWork.tex
There is a rich body of literature in image processing and computer vision on summarizing videos in form of a key frame sequence or a video skim. It is beyond the scope of this paper to do a comprehensive review. Interested readers can check~\cite{money2008video,Truong2007} for a more comprehensive summary. Roughly, all these summarization methods can be divided into two categories: single-video and multi-video summarization.

\vspace{1mm}
\underline{\textit{Single-Video Summarization:}} Much progress has been made in developing a variety of ways to summarize a single video in an unsupervised manner or developing supervised algorithms. Representative methods along the direction of supervised algorithms use category-specific classifiers for importance scoring~\cite{Category2014,Ranking2014} or learn how to select informative and diverse video subsets from human-created summaries~\cite{gygli2015video,gong2014diverse,zhang2016summary} or learn important facets, like faces, hands, objects, diversity~\cite{Graumann2012,lu2013story,Sigal2015}. 
Although these supervised techniques have shown impressive results, their performance largely depends on huge amount of labeled examples which are difficult to collect in many cases. Nevertheless, it is generally feasible to have only a limited number of users to annotate training videos, which may lead to a biased summarization model.

Without supervision, summarization methods rely on low-level visual indices to determine the important parts of a video. Various strategies have been studied, including clustering~\cite{VSUMM2011,Top2014,herranz2010framework,panda2014scalable}, maximal biclique finding~\cite{chu2015video}, interest prediction~\cite{Att2005,LucVanGool2014}, and energy minimization~\cite{Peleg2007,Onlinecontent2012}. Leveraging crawled web images or videos is also another recent trend for video summarization~\cite{Khosla2013,song2015tvsum,Joint2014,panda2017collaborative}.

Recently, there has been a growing interest in using sparse coding (SC) to
solve the problem of video summarization~\cite{Ehsan2012,Eric2014,Scalable2012,meng2016keyframes,dornaika2015decremental} since the sparsity and reconstruction error term in SC naturally fits into the problem of summarization. Another recent work~\cite{elhamifar2014dissimilarity} finds a subset of the source set to efficiently describe the target set, given pairwise dissimilarities between two sets. 
In contrast to these prior works that can only summarize a single video, we develop a multi-video summarization method that jointly summarizes a set of videos to find a single summary for describing the collection altogether. Moreover, we consider interestingness of segments along with representativeness in the sparse optimization to extract summaries that are both interesting and representative.

\vspace{1mm}
\underline{\textit{Multi-Video Summarization:}} Generating a summary from multiple videos is a more technically challenging problem due to the inevitable thematic diversity and content overlaps within multiple videos than a single video. 
Generally, the applications of multi-video summarization can be roughly divided into two categories. The first category is to summarize a group of topically related web videos given by a search. Some of early works in this category focused on videos of specific genres, such as TV news~\cite{li2010multi,wang2009multi} and generated an automatic summary by frame clustering~\cite{yahiaoui2001generating} or leveraging genre specific information, e.g., speech transcripts in news~\cite{li2010multiAV,shao2010multi}. However, they generally fail to summarize large scale open world web videos since they are unstructured and range over a wide variety of content. A system for rapid browsing of multiple videos are proposed in~\cite{dale2012multi}. A recent approach to the problem of summarizing multiple sensor-rich videos in geo-space can be seen in~\cite{zhang2012multi}. A supervised approach to summarize multiple videos captured with hand-held devices is presented in~\cite{zhang2015effective}. However, these systems relies on meta-data sensor information or semantics related to a geographical area (e.g., weather and lighting condition) which are mostly unavailable while summarizing unconstrained web videos.

The other category of multi-video summarization is to summarize videos captured with video sensors at the same time with overlapped or partially overlapped field-of-views in a camera network. Representative methods in this category use random walk over spatio-temporal shot graphs~\cite{MultiviewTMM2010} and rough sets~\cite{MultiviewICIP2011} to summarize multi-view videos. A recent work in~\cite{SanjaySir2015} uses bipartite matching constrained optimum path forest clustering to solve the problem of summarizing multi-view videos. 
An online method for summarization can also be found in~\cite{OnlineMultiview2015}. In~\cite{ManjunathAccv2011,ManjunathTON2014}, summarization is performed by detecting abnormal events between sensors in a non-overlapping camera network. 

Since both of the categories of multi-video summarization are inherently related, we develop, to our best knowledge, the first generalized framework to extract an informative summary by exploring the complementary information within multiple videos. We demonstrate the generalizability of our framework with extensive experiments on several datasets.

%% file: Methodology.tex
In this section, we start by giving notations and definitions of the main concepts of our approach, and then present our detailed approach to summarize multiple videos. 

\vspace{1mm}
\textit{Notation:} We use uppercase letters to denote matrices and lowercase letters to denote vectors. For matrix $A=(a_{ij})$, its $i$-th row and $j$-th column are denoted by $a_i$ and $a^j$ respectively. $||A||_F$ is Frobenius norm of $A$ and $tr(A)$ denote the trace of $A$. The $\ell_p$-norm of the vector $a \in \mathbb{R}^n$ is defined as $||a||_p = (\sum_{i=1}^{n}|a_i|^p)^{1/p}$ and $\ell_0$-norm is defined as $||a||_0 = \sum_{i=1}^{n}|a_i|^0$. The Frobenius norm of $A \in \mathbb{R}^{n\times m}$ is defined as $\sqrt{\sum_{i=1}^{n}\sum_{j=1}^{m}a_{ij}^2}$. The $\ell_{2,1}$-norm can be generalized to $\ell_{r,p}$-norm which is defined as $||A||_{r,p}= (\sum_{i=1}^{n}||a_i||_{r}^{p})^{1/p}$. When $r \geq 1$ and $p \geq 1$, the $\ell_{r,p}$-norm is a valid norm since it satisfies the three basic conditions of a norm including the triangle inequality $||A||_{r,p}+||B||_{r,p} \geq ||A+B||_{r,p}$. However, when $r < 1$ or $p < 1$, $\ell_{r,p}$-norm is not valid as well as the $\ell_0$, but we also call them norms for convenience. The operator $diag(.)$ puts a vector on the main diagonal of a matrix. 1 denotes a vector whose elements are equal to one.        

\vspace{2mm}
\textit{Video Summary:} Given a set of videos, our goal is to find a summary that conveys the most \textit{important} details of the original video collection. Specifically, it is composed of several video segments that represent most important portions of the input video collection within a short duration.    
Since, \textit{importance} is a subjective notion, we define a good summary as one that has the following properties:
\newline
$\bullet$ \textit{Representativeness.} The set of videos should be reconstructed with high accuracy using the extracted summary.
\newline
$\bullet$ \textit{Interestingness.} The summary should contain the most interesting parts of the input videos, e.g., in a collection of videos related to \textit{Eiffel Tower}, one does not want to miss a segment that depicts the colorful night view of the tower. 
\newline
$\bullet$ \textit{Sparsity.} Although the summary should be representative and interesting, the length should be as small as possible.
\newline
$\bullet$ \textit{Diversity.} The summary should be diverse as much as possible capturing different aspects of the input video collection. In other words, the amount of content redundancy should be small in the final set of extracted summaries. 

\vspace{1mm}
We develop a diversity-aware sparse optimization framework to generate a multi-video summary that characterizes all the above desirable properties of an optimal summary. 
The proposed approach, \textsf{DiMS}, decomposes into three steps: i) video representation; ii) diversity-aware sparse representative selection; iii) summary generation.
\vspace{-3mm}
\subsection{Video Representation}
\label{sec:video representation}

Video representation is a crucial step in summarization for maintaining visual coherence, which in turn affects the overall quality of a summary. It basically consists of two main steps, namely, (i) temporal segmentation, and, (ii) feature representation. We describe these steps in the following sections. 
\vspace{1mm}
\subsubsection{\underline{Temporal Segmentation}} 
Our approach starts with segmenting videos using an existing algorithm~\cite{chu2015video}. We divide each video into multiple non-uniform segments by measuring the amount of changes between two consecutive frames in the RGB and HSV color spaces~\cite{boreczky1996comparison}. A segment boundary is determined at a certain frame when the portion of total change is greater than 75\%~\cite{chu2015video}. We added an additional constraint to the segmentation algorithm to ensure that the number of frames within each segment lies in the range of [32,96]. The video segments serve as the basic units for feature extraction and subsequent processing to extract a video summary.
\vspace{1mm}
\subsubsection{\underline{Feature Representation}} Deep convolutional neural networks (CNNs) have been successful at large-scale object recognition~\cite{krizhevsky2012imagenet}.
Beyond the object recognition task itself, recent advancement in deep learning has revealed that features extracted from upper or intermediate layers of a CNN are generic features that have good transfer learning capabilities across different domains~\cite{simonyan2014two,karpathy2014large}.
An advantage of using deep learning features is that there exist accurate, large-scale datasets such as Imagenet~\cite{russakovsky2015imagenet}, and Sports-1M~\cite{karpathy2014large} from which they can be extracted.
In addition, GPU-based extraction of such features are much faster than that for the traditional hand crafted features such as CENTRIST, Dense-SIFT.

In the case where the input is a video clip, C3D features~\cite{tran2014c3d} have recently shown better performance compared to the features extracted using each frame separately~\cite{tran2014c3d}.
We therefore extract C3D features, by taking sets of 16 input frames, applying 3D convolutional filters, and extracting the responses at FC6 layer as suggested in~\cite{tran2014c3d}. 
This is followed by a temporal mean pooling scheme to maintain the local ordering structure within a video segment. 
Then the pooling result serves as the final feature vector of a video segment (4096 dimensional) to be used in the sparse optimization. We will discuss the performance benefits of employing C3D features later in our experiments.

Note that in our current work, we did not consider the audio information while representing videos. However, we believe that audio (if available) can be used as a potential side information along with visual features to select important segments from a video. One can easily incorporate audio features in our framework by combining both audio and visual features to represent a video segment or following aggregation mechanisms similar to~\cite{li2010multiAV,harwath2015deep}---we leave this as an interesting direction for future research. Our proposed sparse optimization approach as described in next section, is quite flexible in handling multi-modal information while summarizing videos---we expect more sophisticated ones will only benefit our approach. 
\vspace{-3mm} 
\subsection{Diversity-aware Sparse Representative Selection}
We develop a sparse optimization framework that jointly summarizes a set of videos to extract a summary that describes the collection together. Consider a set of $m$ relevant videos given by a video search or generated from a multi-view camera network, where ${{X^{\it{(v)}}}} = \{{{x}^{\it i}} \in \mathbb{R}^d ,i = 1,\cdots,n_v\}, v = 1,\cdots,m$. 
Each ${{x}^{\it i}}$ represents the feature descriptor of a segment in $d$-dimensional feature space. We represent each video segment by extracting C3D features as described above. 
\subsubsection{\underline{Formulation}}
Sparse optimization approaches~\cite{Scalable2012,Ehsan2012} find the representative segments from a single video ${{X^{\it{(v)}}}}$ by minimizing the linear reconstruction error as
\begin{equation}
\begin{gathered}
\label{eq:singlevideo0}
\min_{{Z^{(\it{v})}}}\ \lVert {{X^{\it{(v)}}} - {X^{\it{(v)}}Z^{\it{(v)}}} \rVert}^2_{F} 
\ s.t.\  \lVert {{Z^{\it{(v)}}}\rVert}_{2,0} \leq  k, \ {Z^{(\it{v})^{T}}}1 = 1
\end{gathered}
\end{equation}  
The constraint on $\ell_{2,0}$ norm of ${Z^{\it{(v)}}}$ implies that only $k$ video segments are chosen as the representative whereas the affine constraint ${{Z^{(\it{v})^{T}}}1 = 1}$ makes the selection of representatives invariant with respect to the global translation of the data.

This is an NP-hard problem since it requires searching over every subset of the $k$ columns of ${{X^{\it{(v)}}}}$. A standard $\ell_{1}$ relaxation to the problem (\ref{eq:singlevideo0}) is given by
\begin{equation}
\begin{gathered}
\label{eq:singlevideo00}
\min_{{Z^{(\it{v})}}}\ \lVert {{X^{\it{(v)}}} - {X^{\it{(v)}}Z^{\it{(v)}}} \rVert}^2_{F}
\ s.t.\  \lVert {{Z^{\it{(v)}}}\rVert}_{2,1} \leq  \tau, \ {Z^{(\it{v})^{T}}}1 = 1
\end{gathered}
\end{equation}
where ${\lVert {{Z^{\it{(v)}}}}\rVert}_{2,1} = \sum_{i=1}^{n_v}||z^{\it{(v)}}_i||_2$ and $\tau > 0 $ controls the level of sparsity in the reconstruction.\footnote{Note that we use $\tau$ instead of $k$ since  ${\lVert {Z}\rVert}_{2,1}$ is not necessarily bounded by $k$ after the relaxation.}
Using Lagrange multiplier, the optimization problem (\ref{eq:singlevideo00}) can be written as,   
\begin{equation}
\begin{gathered}
\label{eq:singlevideo}
\min_{{Z^{(\it{v})}}}\ \lVert {{X^{\it{(v)}}} - {X^{\it{(v)}}Z^{\it{(v)}}} \rVert}^2_{F} + \lambda^{\it{(v)}}_s\lVert {{Z^{\it{(v)}}}\rVert}_{2,1} 
\ s.t.\ \ {{Z^{(\it{v})^{T}}}1 = 1}
\end{gathered}
\end{equation}  
where $\lambda^{\it{(v)}}_s$ is a regularization parameter. Once problem (\ref{eq:singlevideo}) is solved, the summary is generated by selecting segments whose corresponding $||Z^{\it{(v)}}_i||_2 \neq 0$. We keep the constraint ${{Z^{(\it{v})^{T}}}1 = 1}$ since it can be easily handled as we will show later. 

\vspace{1mm}
\underline{\textit{Introducing Interestingness of Video Segments:}} Note that in problem (\ref{eq:singlevideo}), all segments are treated equally without considering the interestingness of some specific segments. Specifically, sparse optimization approaches~\cite{Scalable2012,Ehsan2012} only characterizes the reconstruction capability and sparsity but does not account for the fact that the selected video segments should be interesting. As a result, it may leave out some crucial segment(s) in the summary. A good summarization method can certainly benefit from incorporating such interestingness prior knowledge from application domain or user specifications. To better leverage interestingness along with representativeness, we propose a simple extension to (\ref{eq:singlevideo}) as follows~\cite{meng2016keyframes}:
\begin{equation}
\begin{gathered}
\label{eq:singlevideoInt}
\min_{{Z^{(\it{v})}}}\ \lVert {{X^{\it{(v)}}} - {X^{\it{(v)}}Z^{\it{(v)}}} \rVert}^2_{F} + \lambda^{\it{(v)}}_s\lVert Q^{(v)}{{Z^{\it{(v)}}}\rVert}_{2,1} 
\ s.t.\ \ {{Z^{(\it{v})^{T}}}1 = 1}
\end{gathered}
\end{equation}  
where $Q^{(v)}=[diag(q^v)]^{-1}$ and $q^v \in \mathbb{R}^{n_v}$ represent the interstingness score of each video segment. It is easy to see that problem (\ref{eq:singlevideoInt}) favors selection of interesting segments by assigning a lower score via $Q^{(v)}$. Thus, given a video, minimization of (\ref{eq:singlevideoInt}) selects segments that are both interesting and representative. More details on the video interestingness prior are presented in Sec.~\ref{sec:Discussions}.  

The sparse optimization (\ref{eq:singlevideoInt}) extracts a good summary from a single video. However, summarizing multiple videos is ubiquitous in video search or in a camera network, hence, extending (\ref{eq:singlevideo}) into multi-video setting is of vital importance for many multimedia applications. One direct way to extend into multi-video setting is to apply (\ref{eq:singlevideo}) to each of the video, and then combine the results to produce a single summary. Mathematically, we have the na{\"i}ve multi-video summarization approach as follows:
\vspace{-3mm}
\begin{equation}
\begin{gathered}
\label{eq:naivemultivideo}
\min_{{Z^{(1)}},{Z^{(2)}},\cdots,{Z^{({m})}}} \ \sum_{v=1}^{m}\lVert {{X^{\it{(v)}}} - {X^{\it{(v)}}Z^{\it{(v)}}} \rVert}^2_{F} + \sum_{v=1}^{m} \lambda^{\it{(v)}}_s\lVert Q^{(v)} {{Z^{\it{(v)}}}\rVert}_{2,1} \\
s.t.\ \ {{Z^{(\it{v})^{T}}}1 = 1},\ {{Z^{(\it{v})}}} \in \mathbb{R}^{n_v\times n_v},\ \forall \ 1 \leq \it{v} \leq m	
\end{gathered}
\end{equation}
This approach summarizes videos independently without considering complementarity of different videos, hence, produces redundant information in the final summary.
         
\vspace{1mm}  
\underline{\textit{Introducing Complementarity of Multiple Videos:}}
The objective function (\ref{eq:naivemultivideo}) summarizes multiple videos independently, without any constraint.
Considering the presence of complementary information within multiple videos, we introduce a diversity regularization function to select a sparse set of representative and diverse video segments. Specifically, to explore the complementary information, we enforce a regularizer that penalizes the condition that two correlated segments from two distinct videos are present in the summary at the same time. For example, if the $i$-th segment from $v$-th video is highly correlated to the $j$-th segment in $w$-th video, then we do not need to select both of them simultaneously.

\vspace{1mm}
\textit{{Definition 1.} Given the sparse coefficient matrices ${{Z^{(\it{v})}}}$ and ${{Z^{(\it{w})}}}$, the diversity regularization function is defined as:}
\vspace{-2mm}   
\begin{equation}
\begin{gathered}
\label{eq:diversity}
{f_{d}({{Z^{(\it{v})}}},{Z^{(\it{w})}})} = \sum_{i=1}^{n_v}\sum_{j=1}^{n_w}\lVert{{z^{(\it{v})}_{\it{i}}} \rVert}_{2}{{c_{\it{ij}}}}\lVert{{z^{(\it{w})}_{\it{j}}} \rVert}_{2} 
=\lVert{{W^{\it{(vw)}}}{Z^{(\it{v})}} \rVert}_{2,1}
\end{gathered}
\end{equation}  
where ${{c_{\it{ij}}}}$ measure the correlation between $i$-th segment from $v$-th video and the $j$-th segment in $w$-th video.
The second equality follows from the simple manipulation as ${{W^{\it{(vw)}}_{\it{ii}}}} = \sum_{j=1}^{n_w}{c_{\it{ij}}}\lVert{{z^{(\it{w})}_{\it{j}}}\rVert}_{2,1}$. More details on correlation between different video segments are given in Sec.~\ref{sec:Discussions}.  

\vspace{1mm}
Minimization of (\ref{eq:diversity}) tries to explore the complementarity by penalizing the condition that rows of two similar video segments from two distinct videos are nonzero at the same time. 
This amounts to enforcing the sparse coefficient matrices of different videos to be of maximum diversity.

\vspace{1mm}
\underline{\textit{Overall Objective Function:}} After adding the diversity regularization function into problem (\ref{eq:naivemultivideo}), we have the final objective function as follows:
\vspace{-1mm}    
\begin{equation}
\begin{gathered}
\label{eq:multivideo}
\min_{{Z^{(1)}},{Z^{(2)}},\cdots,{Z^{({m})}}} \sum_{v=1}^{m}\lVert {{X^{\it{(v)}}} - {X^{\it{(v)}}Z^{\it{(v)}}} \rVert}^2_{F} + \lambda_s \sum_{v=1}^{m} \lVert Q^{(v)}{{Z^{\it{(v)}}}\rVert}_{2,1} \\ + \lambda_d \sum_{\substack{1\leq v,w\leq m \\ v \neq w}}f_{d}({{Z^{(\it{v})}}},{Z^{(\it{w})}}) \\  
s.t.\ \ {{Z^{(\it{v})^{T}}}1 = 1},\ {{Z^{(\it{v})}}} \in \mathbb{R}^{n_v\times n_v},\ \forall \ 1 \leq \it{v} \leq m	
\end{gathered}
\end{equation}
where $\lambda_s$ and $\lambda_d$ are two trade-offs associated with the sparsity and diversity regularization functions respectively.

\vspace{1mm}
\subsubsection{\underline{Optimization}} It is difficult to solve the constrained problem (\ref{eq:multivideo}). In this section, we propose an alternative algorithm to solve this optimization problem efficiently. 
With the alternating minimizing strategy, we can approximately solve (\ref{eq:multivideo}) in the manner of minimizing with respect to one video once at a time while fixing the other videos.
Specifically, we minimize the following objective function with respect to  ${Z^{(\it{v})}}$ while keeping all others fixed:
\begin{equation}
\begin{gathered}
\label{eq:multivideowithonev}
\min_{{Z^{(\it{v})}}}\ \lVert {{X^{\it{(v)}}} - {X^{\it{(v)}}Z^{\it{(v)}}} \rVert}^2_{F} + \lambda_s \lVert Q^{(v)}{{Z^{\it{(v)}}}\rVert}_{2,1}  \\ +  {\lambda_d \sum_{w=1, v \neq w}^{m}\lVert{{W^{\it{(vw)}}}{Z^{(\it{v})}} \rVert}_{2,1}}  
\ s.t. \ \ {{Z^{(\it{v})^{T}}}1 = 1}
\end{gathered}
\end{equation}
To reformulate the problem (\ref{eq:multivideowithonev}), we need the following lemma.

\vspace{1mm}
\textit{{Lemma 1.} For any two diagonal positive semidefinite matrices ${{W^{(\it{1})}}},{{W^{(\it{2})}}} \in \mathbb{R}^{n\times n}$, the following equality holds for any matrix ${{Z}} \in \mathbb{R}^{n\times n}$:}
 \begin{equation}
 \begin{gathered}
 \label{eq:lemma}
 {\lVert{{W^{\it{(1)}}}{Z}} \rVert}_{2,1} + {\lVert{{W^{\it{(2)}}}{Z}}\rVert}_{2,1}={\lVert{{W}{Z}} \rVert}_{2,1}
 \end{gathered}
 \end{equation}
 where ${W}={W}^{\it{(1)}}+{W}^{\it{(2)}}$. The proof follows directly from the fact that $\ell_{2,1}$-norm is a valid norm and the equality in triangle inequality holds if both ${{W^{\it{(1)}}}}$ and ${{W^{\it{(2)}}}}$ are positive semidefinite matrices.    $\square$ 

\vspace{1mm}
From lemma 1, it is easy to reformulate problem (\ref{eq:multivideowithonev}) as following:
\begin{equation}
\begin{gathered}
\label{eq:onev1}
\min_{{Z^{(\it{v})}}}\ \lVert {{X^{\it{(v)}}} - {X^{\it{(v)}}Z^{\it{(v)}}} \rVert}^2_{F} + \lambda_s \lVert Q^{(v)}{{Z^{\it{(v)}}}\rVert}_{2,1} + \lambda_d \lVert{{W^{\it{(v)}}}{Z^{(\it{v})}} \rVert}_{2,1} \\  
s.t.\ \ {{Z^{(\it{v})^{T}}}1 = 1}
\end{gathered}
\end{equation}
where ${{W^{\it{(v)}}}} = \sum_{w=1, v \neq w}^{m}{{W^{\it{(vw)}}}}$. Note that both second and third term in (\ref{eq:onev1}) are functions of the same variable ${Z^{(\it{v})}}$ with two trade-offs $\lambda_s$ and $\lambda_d$ respectively. From lemma 1, we can approximate (\ref{eq:onev1}) with one trade-off parameter $\lambda$ as following: 
\begin{equation}
\begin{gathered}
\label{eq:onev3}
\min_{{Z^{(\it{v})}}}\ \lVert {{X^{\it{(v)}}} - {X^{\it{(v)}}Z^{\it{(v)}}} \rVert}^2_{F} + \lambda \lVert {{K^{\it{(v)}}}{Z^{\it{(v)}}}\rVert}_{2,1}   
\ s.t.\ \ {{Z^{(\it{v})^{T}}}1 = 1}	
\end{gathered}
\end{equation}
where ${K^{\it{(v)}}}={Q^{(v)}+W^{\it{(v)}}}$. For convenience, ignoring the superscripts, we get
\begin{equation}
\begin{gathered}
\label{eq:onevfinal2}
\min_{{Z}}\ \lVert {{X} - {X}Z \rVert}^2_{F} + \lambda \lVert {{Z}\rVert}_{{K},2,1} \  
s.t.\ \ {{Z^{\it T}}1 = 1}
\end{gathered}  
\end{equation}
where $\lVert {{Z}\rVert}_{{K},2,1}$ denotes the weighted $\ell_{2,1}$-norm of ${Z}$ and is defined as $\lVert {{Z}\rVert}_{{K},2,1}=\lVert { K{Z}\rVert}_{2,1}$. When we replace ${X}$ with $[{X}^{\it T},\alpha*{1}]^{\it T}$ where $\alpha$ approaches to infinity, (\ref{eq:onevfinal2}) is equivalent to the following problem:
\begin{equation}
\begin{gathered}
\label{eq:onevfinal3}
\min_{{Z}}\ \lVert {{X} - {X}Z \rVert}^2_{F} + \lambda \lVert {{Z}\rVert}_{{K},2,1}
\end{gathered}
\end{equation}
We can prove equation (\ref{eq:onevfinal2}) is equivalent to (\ref{eq:onevfinal3}) by expanding (\ref{eq:onevfinal3}) as follows:
\begin{equation}
\begin{gathered}
\label{eq:onevfinal4}
\lVert {{X} - {XZ} \rVert}^2_{F}={\lVert {{X}^{\it{*}}} - {X}^{\it{*}}Z \rVert}^2_{F}+\alpha {\lVert {1^{\it{T}}}-{1^{\it{T}}}{Z} \rVert}^2_{F}
\end{gathered}
\end{equation}
where ${{X}^{\it{*}}}$ is the original ${{X}}$ presented in (\ref{eq:onevfinal2}). When $\alpha$ approaches to infinity, ${{Z^{\it T}}1}$ approaches to ${1}$. Thus, problem (\ref{eq:onevfinal2}) is equivalent to (\ref{eq:onevfinal3}).

The objective function (\ref{eq:onevfinal3}) is a convex weighted $\ell_{2,1}$-norm minimization problem which can be efficiently solved using Alternating Direction Method of Multipliers (ADMM)~\cite{boyd2011distributed}. The ADMM procedure to solve (\ref{eq:onevfinal3}) is summarized in Algo.~\ref{algo:ADMM}.
\footnote{We provide details about the ADMM in the supplementary material. The supplementary material associated with this paper is available at \url{http://www.ee.ucr.edu/~amitrc/publications.php}}

The above alternating procedure of \textsf{DiMS} is carried out until convergence, as shown in Algo.~\ref{algo:OverallALM}.      
\begin{algorithm} 
	\caption{An ADMM solver for (\ref{eq:onevfinal3})}\label{algo:ADMM}
	\begin{algorithmic}
		\State {\bf Input:} Video feature matrix ${X}$, ${K}$ and $\lambda, \mu >0$  
		\While{\textit{not converged}} \\
		\ \ ${U} \leftarrow ({X}^{T}{X}+\mu{I})^{-1}({X}^{T}{X}+\mu{Z}-\mathbf{\Lambda})$;\\
		\ \ ${Z} \leftarrow \max  \Big\{{\lVert {U}+{\Lambda}/\mu \rVert}_2 - \frac{\lambda{K}}{\mu},0\Big\}\frac{{U}+{\Lambda}/\mu}{{\lVert {U}+{\Lambda}/\mu \rVert}_2}$ (row-wise);\\
		\ \ ${\Lambda} \leftarrow {\Lambda} + \mu({U-Z})$;
		\EndWhile
		\State {\bf Output:} Sparse coefficient matrix ${Z}$.
	\end{algorithmic}
\end{algorithm}
\vspace{-5mm}
\begin{algorithm} 
	\caption{Algorithm for solving (\ref{eq:multivideo})}\label{algo:OverallALM}
	\begin{algorithmic}
		\State {\bf Input:} Video feature matrices ${{X^{(\it 1)}},{X^{(\it 2)}},\cdots,{X^{({\it m})}}}$ 
		\For{\textit{each v}} \\
		\ \ Initialize ${{Z^{(\it{v})}}}$ by solving (\ref{eq:naivemultivideo});
		\EndFor
		\While{\textit{not converged}} 
		\ \ \For{\textit{each v}} \\
		\ \ \ \ \ \ \  Obtain ${{Z^{(\it{v})}}}$ by solving (\ref{eq:onevfinal3}); 
		\EndFor
		\EndWhile
		\State {\bf Output:} Coefficient matrices ${{Z^{(1)}},{Z^{(2)}},\cdots,{Z^{({m})}}}$.
	\end{algorithmic}
\end{algorithm}          

\vspace{-5mm} 
\subsection{Summary Generation}
\label{sec:Summary Generation}
Above, we described how we compute the sparse coefficient matrices where the nonzero rows indicate the representatives for the summary.
We follow the following rules to generate a summary of specified length:
(i) We first sort the representative segments in a video ${{X^{\it (v)}}}$ by decreasing importance according to the $\ell_{2}$ norms of the rows in ${{Z^{(\it v)}}}$ (resolving ties by favoring shorter video segments).
(ii) We then sort the videos according to the number of nonzero rows in the corresponding sparse coefficient matrix (informative score) and compute the number of segments that should be selected from each video based on the relative score and user-defined summary length.
(iii) Finally, we construct the video summary by placing the selected segments from the most informative video at the beginning and then appending segments from other videos based on the relative informative score.

%% file: Discussions.tex
\underline{\textit{Interestingness of Video Segments:}} As existing approaches, we compute the interestingness score of each segment by taking into account the rest of the video segments. Specifically, we first compute the interestingness score of a segment as the sum of scores predicted for each frame that belong to the segment and then take the relative score over the maximum predicted score in a video. We follow~\cite{LucVanGool2014} to compute the interestingness score of each frame by considering attention, aesthetic quality and presence of landmarks/persons. Note that these forms of interestingness prediction are often used in several vision tasks and are quite flexible~\cite{Graumann2012,ejaz2013efficient,datta2006studying,dhar2011high}. However, one can also learn a regression model to predict an interestingness score of domain relevance~\cite{gygli2015video,Ranking2014,yaohighlight} or compute with user specifications via human in the loop~\cite{han2011personalized}---we expect more sophisticated ones will only benefit our proposed approach. Concretely speaking, our method is not dependent on a particular definition on interestingness. 

\vspace{1mm}
\underline{\textit{Correlation between Video Segments:}} There are a lot of ways to measure the correlation between two video segments ${{c_{\it{ij}}}}$. In this paper, we employ Scott and Longuet-Higgins (SLH) algorithm~\cite{Scott1991} with Gaussian kernel to measure the correlation, since it is simple to implement and it performs well in several vision tasks~\cite{permutation2013,torki2010one}.
Specifically, given the segment-level feature similarity matrix ${S}$, computed via a Gaussian kernel between two videos, SLH algorithm finds an orthonormal matrix ${C}$ that permutes the rows of $S$ in order to maximize its trace. Mathematically,
\vspace{-1mm}
\begin{equation}
\begin{gathered}
\label{eq:SLH}
{C} = \argmax_{{C}^T {C}=I} \  tr({C}^T {S})
\end{gathered}
\vspace{-1mm}
\end{equation} 
Maximizing the above function is a singular value decomposition problem and the optimal solution is given by ${C}^{*} = {U}{D}{V}^T$, where the SVD decomposition of ${S}={U}{E}{U}^T$ and ${D}$ is obtained by replacing singular values of ${E}$ by ones. We use the matrix ${C}^{*}$ as the correlation matrix after setting the negative values to 0~\cite{torki2010one}. 
We will discuss the performance benefits in employing such correlations compared to the cosine similarity in the experiments. It is also important to mention here that our proposed formulation (\ref{eq:multivideo}) is highly flexible to incorporate any form of correlations defined between two video segments.       

\vspace{1mm}
\underline{\textit{Sparsity Regularization Parameter:}} The regularization parameter $\lambda$ in (\ref{eq:onevfinal3}) puts a trade-off between two opposing terms: the reconstruction error and number of representative segments. In other words, we obtain a small reconstruction error by selecting more representative segments and vice versa. As indicated by the update equation of $Z$ in Algo.~\ref{algo:ADMM}, when $\lambda$ is large enough, e.g., $\lambda \geq \lambda^{max}$, we get $Z=0$ that means we select no representative segments. 
Thus, to avoid an empty selection, we let $\lambda \leq \lambda^{max}$ and obtain $\lambda^{max}= \max_{0 \leq i \leq n} ||x_i^{T}X||_2$, as in~\cite{Ehsan2012}.
In our experiments, we let $\lambda=\frac{\lambda^{max}}{\alpha}$ and tune $\alpha$ between the interval [2,30]~\cite{Ehsan2012}. 

\vspace{1mm}
\underline{\textit{Intialization in Algo.~\ref{algo:OverallALM}:}} Since the alternating minimization can make the Algo.~\ref{algo:OverallALM} stuck in a local minimum, it is important to have a sensible initialization.
We initialize the sparse coefficient matrices of $m-1$ videos by solving (\ref{eq:naivemultivideo}) using Algo.~\ref{algo:ADMM}, which is a special case (when $\lambda_d = 0$ in (\ref{eq:multivideo})) of our method. After the initialization, the following question remain: from which view we should start the alternating minimization? One possible way is to randomly start with any video and repeat the minimization over all videos until convergence. However, since we have some prior knowledge on which video is more informative in the collection, we can start with initializing and fixing more informative videos, and optimize with respect to the least informative video. More specifically, we start with the specific ${{Z^{(\it{v})}}}$ which has more number of nonzero rows after solving (\ref{eq:naivemultivideo}) since the number of nonzero rows indicate the relative importance of each video in the collection.

\vspace{1mm}
\underline{\textit{Stopping Criteria:}} In Algo.~\ref{algo:ADMM}, the stop criteria is set to $||U^{(t)}-Z^{(t)}||_{\infty} \leq \epsilon  \ or \ t \geq 2000$, where $t$ is the iteration number and $\epsilon$ is set to $10^{-7}$ throughout the experiments. Similarly, in Algo.~\ref{algo:OverallALM}, we set the stop criteria as $\frac{|f^{(t+1)}-f^{(t)}|}{f^{(t)}}<10^{-2}$, where ${f^{(t)}}$ is the objective value in the $t$-th iteration.

\vspace{1mm}
\underline{\textit{Convergence Analysis:}} We can prove the convergence of the proposed Algo.~\ref{algo:OverallALM} as follows: we divide the problem (\ref{eq:multivideo}) into $m$ number of subproblems and each of them is a convex problem with respect to one variable (Algo.~\ref{algo:ADMM}). The convergence of Algo.~\ref{algo:ADMM} is guaranteed by the existing Alternating Direction Method of Multipliers (ADMM) theory~\cite{glowinski1989augmented}. Therefore, by solving the subproblems alternatively, our proposed algorithm will guarantee that we can find the optimal solution to each subproblem and finally, the algorithm will converge to the local solution. 
In all our experiments, we monitor the convergence is reached within less than 10 iterations.

\vspace{1mm}
\underline{\textit{Time Complexity Analysis:}} 
As discussed earlier, our overall problem can be divided into $m$ number of subproblems and each of them can be solved using Algo.~\ref{algo:ADMM}, we first analyze the computational complexity of Algo.~\ref{algo:ADMM} and then present the total complexity of our method. We also show that the proposed approach allows for parallel implementation, which can further reduce the computational time to a large extent. 

In Algo.~\ref{algo:ADMM}, each iteration contains three substeps (See Appendix for details):
(i) solving a linear system with respect to $U$ for once and is not repeated for each iteration. Solving this requires at most complexity of $\mathcal{O}(n_v^{3})$. However, we can solve this via $n_v$ independent smaller linear systems over the $n_v$ columns of $U$. Thus, with $P$ parallel processing resources, we can reduce the computational time to $\mathcal{O}(n_v^{3}/P)$,
(ii) update with respect to $Z$ can be done in $\mathcal{O}(n_v^{2})$ computational time. However, since the solution correspond to one-dimensional shrinkage and thresholding operation, we can perform the update via $n_v$ indepedent shrinkage operations over the $n_v$ rows of $Z$. Thus, with $P$ parallel processing resources, this can be reduced to $\mathcal{O}(n_v^{2}/P)$,
(iii) similarly, update on $\Lambda$ can be done in $\mathcal{O}(n_v^{2}/P)$ computational time with $P$ parallel processing resources by performing $n_v$ independent updates over rows or columns. As a result, the computational complexity of Algo.~\ref{algo:ADMM} is $\mathcal{O}(n_v^{3}+2*n_v^{2}) \approx \mathcal{O}(n_v^{3})$ and it reduces to $\mathcal{O}(n_v^{3}/P)$ with $P$ parallel processing resources. The proposed approach invokes Algo.~\ref{algo:ADMM} for each subproblem i.e., with respect to one video alternatively. By adopting the same procedure, the computational complexity of our approach is $\mathcal{O}(\sum_{v=1}^{m}{n_v^{3}})$. Note that time complexity for solving a linear system can be reduced from $\mathcal{O}(n_v^{3})$ to $\mathcal{O}(n_v^{2.376})$ using the Coppersmith-Winograd algorithm. Therefore, the time complexity of our approach is $\mathcal{O}(\sum_{v=1}^{m}{n_v^{2.376}})$ and it reduces to $\mathcal{O}([\sum_{v=1}^{m}{n_v^{2.376}}]/P)$ with $P$ parallel processing resources.

%% file: Experiments.tex
In this section, we present various experiments and comparisons
to validate the effectiveness and efficiency of our proposed algorithm in two summarization tasks such as topic-oriented video summarization and multi-view video summarization in a camera network, as explained below.
\vspace{-4mm}
\subsection{Topic-oriented Video Summarization}
\label{sec:topic}
\underline{\textit{Goal:}} Large collections of web videos contain clusters of videos belonging to a topic with typical visual content and repeating patterns across the videos. \textit{Given a set of topic-related videos generated from a video search, can we generate a single summary that describes the collection altogether?} Specifically, our goal is to generate a single video summary that can describe the whole video collection. 

\vspace{1mm}
\underline{\textit{Dataset:}} To evaluate topic-oriented video summarization, we need a single ground truth summary of all the topic-related videos that can describe the videos altogether. However, since there exists no such publicly available dataset that fits our need, we introduce a new large dataset, Tour20, that allows for the automatic evaluation of summarization methods in a fast and repeatable manner. We selected 20 tourist attractions from the Tripadvisor travelers choice landmarks 2015 list\footnote{https://www.tripadvisor.com/TravelersChoice-Landmarks\#1}, 
and collected 140 videos from YouTube under the Creative Commons license (See Tab.~\ref{tab:CoSum} for names of the tourist attractions). Such a summary can be a great source of information for prospective tourists when they plan to visit the place and would like to get a preview of its main parts\footnote{Although we focus on summarizing multiple videos of a tourist
attraction as an application area in our experiments, our approach is quite general to summarize any type of videos generated from a search.}.
It is also important to note that all prior works~\cite{zhang2012multi,zhang2015effective,li2010multi,li2010multiAV} conducted experiments on personal test sets, which are not publicly available, thus making it hard for others to reproduce or to compare the presented results. We hope the release of our Tour20 dataset will give researchers a new, dynamic tool to evaluate their video summarization algorithms in a repeatable and efficient way\footnote{The Tour20 dataset along with the complete groundtruth summaries are publicly available to download in \url{http://www.ee.ucr.edu/~amitrc/datasets.php}.}.
To the best of our knowledge, this is the biggest publicly available summarization dataset with 140 videos totaling about 7 hours (669,497 frames and 12,499 segments). 

\vspace{1mm}
\underline{\textit{Performance Measures.}} Motivated by~\cite{LucVanGool2014,zhang2016summary,song2015tvsum}, we assess the quality of an automatically generated summary by comparing it to human judgment. Specifically, given a proposed summary and a set of human selected summaries, we compute the pairwise F-measure and then report the mean value motivated by the fact that there exists not a single ground truth summary, but multiple summaries are possible. 

\begin{table*} [t]
	\tiny
	\caption{Comparison with Single-video summarization methods on Tour20 dataset. Numbers show mean F-measures at 10\% summary length, \textit{i.e.}, summary containing only 10\% of total video segments. We highlight the \textbf{best} and \underline{second best} baseline method. Our approach (\textsf{DiMS}) statistically outperforms all baseline methods by a significant margin $(p < .01)$.} 
	\label{tab:CoSum1}
	\vspace{-2mm}
	\begin{tabulary}{1.1\linewidth}{|p{12mm}||P{13.5mm}|P{13.5mm}|P{13.5mm}|P{13.5mm}|P{13.5mm}|P{13.5mm}|P{8.5mm}|P{8.5mm}|P{9.5mm}|P{12.5mm}|}
		\hline
		\textbf{F-measure} & \textsf{ConcateKmeans}  & \textsf{ConcateSpectral}  & \textsf{ConcateSparse} & \textsf{KmeansConcate} & \textsf{SpectralConcate} & \textsf{SparseConcate} & \textsf{Graph} & \textsf{DT} & \textsf{SubMod} & \textsf{DiMS(ours)} \\
		\hhline{|-|-|-|-|-|-|-|-|-|-|-|}
		\textbf{mean} & 0.396	& 0.413  & 0.450 & 0.455 & 0.465 & 0.503 & 0.457 & 0.476 & \underline{0.512} & \textbf{0.613}\\
		\hline 		
	\end{tabulary} \vspace{-2mm}
\end{table*} 

\vspace{1mm}
\underline{\textit{Ground truth Summaries.}} Previous topic-oriented video summarization approaches generated video summaries and then let humans assess their quality by comparing different system generated summaries. 
Specifically, users are shown different summaries and are asked to select the better one or assign a rating from a predefined scale. 
While simple and fast, this approach does not scale well because the user study has to be re-run every time a change is made. Another alternative is to let the humans watch the whole video and select some of the important segments as the summary. This approach has the advantage that, once the ground truth summaries are obtained, experiments can be carried out indefinitely, which is desirable especially for multimedia systems that involve multiple iterations and testing. We take this approach in our work to generate ground truth summaries.

Given the videos that were pre-processed into several segments, we asked three study experts to select at least 5\%, but no more than 15\% segments for each video as well as a single set of diverse segments that can describe the video collection altogether.
We muted audio to ensure that important video segments are selected based solely on visual stimuli. Moreover, we also specify that if some embedded text is only mentioned in on-screen text, then it should not be labeled as important. 
They could use a simple interface that allows to watch all the videos of a collection at the same time and select important segments from each video.
Note that obtaining these ground truth summaries was very time consuming. The study experts are requested to watch the whole video before selecting ground truth segments as whether a segment is important or not is a relative judgment within a video. Since the dataset contains important segments for each video as well as a diverse set of segments to describe the collection altogether, it can be used to evaluate both single-video and multi-video summarization algorithms in an repeatable and efficient way.

To assert the consistency of human created summaries, we compute both pairwise F-measure and the Cronbach's alpha between them, as in~\cite{LucVanGool2014,song2015tvsum}. The dataset has a mean F-measure of 0.643 and mean Cronobach's alpha of 0.944. Ideally alpha is around 0.9 for a good test~\cite{alpharef}. More details on the dataset consistency and exemplar human created summaries can be found in the supplementary material.      

\vspace{1mm}
\underline{\textit{Compared Methods.}} We compare our approach with several methods that fall into four main categories:
(1) classical clustering based methods such as \textsf{ConcateK means}~\cite{arthur2007k}, \textsf{ConcateSpectral}~\cite{von2007tutorial}, \textsf{ConcateSparse}~\cite{Ehsan2012}, \textsf{KmeansConcate}~\cite{arthur2007k}, \textsf{SpectralConcate}~\cite{von2007tutorial}, \textsf{SparseConcate}~\cite{Ehsan2012}) that use single-video summarization approach over multiple videos to generate a summary. The first three baselines (\textsf{ConcateKmeans, ConcateSpectral, ConcateSparse}) concatenate all the videos into a single video and then apply $k$-means, spectral clustering and sparse coding~\cite{Ehsan2012} to the concatenated video respectively, whereas in the other three baselines (\textsf{KmeansConcate, SpectralConcate, SparseConcate}), the corresponding approach is first applied to each video and then the resulting summaries are combined to form a single video summary.
(2) graph clustering based methods including \textsf{Graph}~\cite{peng2006clip} and \textsf{DT}~\cite{mundur2006keyframe}. \textsf{Graph} uses normalized cut-based clustering~\cite{peng2006clip} over the graph constructed using the concatenated video~\cite{lu2004video}, whereas \textsf{DT} uses Delaunay triangulation-based graph clustering to automatically extract informative and diverse segments from a video. Specifically, a Delaunay graph is first constructed using the video segments and then all the edges are classified into short edges and separating edges using average and standard deviation of edge lengths at each vertex. More details about the Delaunay graph clustering for summarizing videos can be seen in~\cite{mundur2006keyframe}.
We apply Delaunay graph clustering to each video separately and then the resulting summaries are combined to form a single summary.
(3) a submodularity based method (\textsf{SubMod})~\cite{chakraborty2015towards,lin2011class} that uses three selection criteria (Exhaustive, Mutually Exclusive and Interestingness) to extract informative segments from a video. We follow~\cite{chakraborty2015towards} to model the first two selection criteria and follow~\cite{gygli2015video} to model interestingness in summarization. We use the same method~\cite{LucVanGool2014} to compute the interestingness score of each video segment and use a greedy algorithm proposed by Nemhauser \textit{et.al.}~\cite{nemhauser1978analysis} to solve the combined submodular function. Similar to the \textsf{DT} baseline, we apply submodular maximization to each video separately and then the resulting summaries are combined to form a single summary.
(4) state-of-the-art methods including \textsf{MultiVideoContent}~\cite{wang2009multi}, \textsf{MultiVideoMMR}~\cite{li2010multi} which are specifically designed for multi-video summarization. \textsf{MultiVideoContent}~\cite{wang2009multi} uses a greedy approach with a content inclusion measure to summarize multiple videos whereas \textsf{MultiVideoMMR}~\cite{li2010multi} extends the concept of maximal marginal relevance~\cite{carbonell1998use} to the video domain for the same purpose. 

Note that Eq.~(\ref{eq:naivemultivideo}) represents the \textsf{SparseConcate} baseline that summarizes multiple videos without any diversity constraint.
The purpose of comparing with single-video summarization methods is to show that techniques that attempt to find informative summary from a single-video usually do not produce an optimal set of representatives while summarizing multiple videos. Note that the recent two multi-video summarization methods in~\cite{zhang2015effective,zhang2012multi} use meta-data sensor information or semantics related to a geographical area (e.g., weather and lighting condition) and are hence left out for comparison. 

\vspace{1mm}
\underline{\textit{Experimental Settings.}} All methods use the same C3D feature as described in~Sec. \ref{sec:video representation}. For all the compared methods (including ours), we generate a summary at 10\% summary length, i.e., summary containing 10\% of total segments in a video collection. Such a setting can give a fair comparison for various methods. We follow~\cite{zhang2016summary} and utilize VSUMM evaluation  package~\cite{VSUMM2011} for finding matching pair of segments. 

\begin{table} [h]
	\tiny
	\caption{Comparison with Multi-video summarization methods on Tour20 dataset. Numbers show mean F-measures at 10\% summary length. We highlight the \textbf{best} and \underline{second best} baseline method. Overall, our approach (\textsf{DiMS}) statistically significantly outperforms both methods $(p < .01)$. Name of the tourist places are presented in the format \enquote{name (\# videos)}.} 
	\label{tab:CoSum}
	\vspace{-2mm}
	\begin{tabulary}{1.1\linewidth}{|p{30mm}||P{14mm}|P{13mm}|P{11mm}|}
		\hline
		\textbf{Tourist Attractions} & \textsf{MultiVideoContent} & \textsf{MultiVideoMMR} & \textsf{DiMS(ours)}\\
		\hhline{|=|=|=|=|}
		Angkor Wat (7)	&0.431	&\underline{0.452}	&\textbf{0.567}	\\
		Machu Picchu (7)	&0.438	&\underline{0.507}	&\textbf{0.582} \\
		Taj Mahal (7) 	&\underline{0.593}	&0.533	&\textbf{0.679} \\
		Basilica of Sagrada Familia (6) 	&0.488	&\underline{0.492}	&\textbf{0.597}\\
		St. Peter's Basilica (5) 	&0.586	&\underline{0.602}	&\textbf{0.699}\\
		Milan Cathedral (10)	& \underline{0.481}	&0.473	&\textbf{0.571}\\
		Alcatraz (6) 	&0.652	& \underline{0.668}	&\textbf{0.755} \\
		Golden Gate Bridge (6) 		&\underline{0.527}	&0.515	&\textbf{0.618} \\
		Eiffel Tower (8) 	&0.436	& \underline{0.446}	&\textbf{0.562}\\
		Notre Dame Cathedral (8) 	&0.463	&\underline{0.473}	&\textbf{0.550}\\
		The Alhambra (6) 	&0.553	&\underline{0.582}	&\textbf{0.662}\\
		Hagia Sophia Museum (6) 	&0.473	&\underline{0.536} &\textbf{0.585}\\
		Charles Bridge (6)	&0.453	&\textbf{0.534}	&\underline{0.525}\\
		Great Wall at Mutiantu (5) 	&0.493	&\underline{0.507}	&\textbf{0.673} \\
		Burj Khalifa (9) 	&\textbf{0.450}	&0.392	&\underline{0.441} \\
		Wat Pho (5) 	&\underline{0.625}	&0.603	&\textbf{0.722}\\
		Chichen Itza (8) 	&\underline{0.514}	&0.492	&\textbf{0.582}\\
		Sydney Opera House (10) 	&0.503	&\underline{0.512}	&\textbf{0.614}\\
		Petronas Twin Towers (9) 	&0.453	&\underline{0.486}	&\textbf{0.643}\\
		Panama Canal (6) 	&0.512	&\underline{0.544}	&\textbf{0.639} \\
		\hline 		
	\end{tabulary} 	
	\begin{tabulary}{1.1\linewidth}{|p{30mm}||P{14mm}|P{13mm}|P{11mm}|}
		\hline
		\textbf{mean} & \bf{0.506} & \bf{0.517} & \bf{0.613}\\
		\hline 		
	\end{tabulary} 
	\vspace{-3mm}
\end{table}

\begin{figure*} 
	\centering
	\begin{tabular}{c}
		\includegraphics[width=1\linewidth]{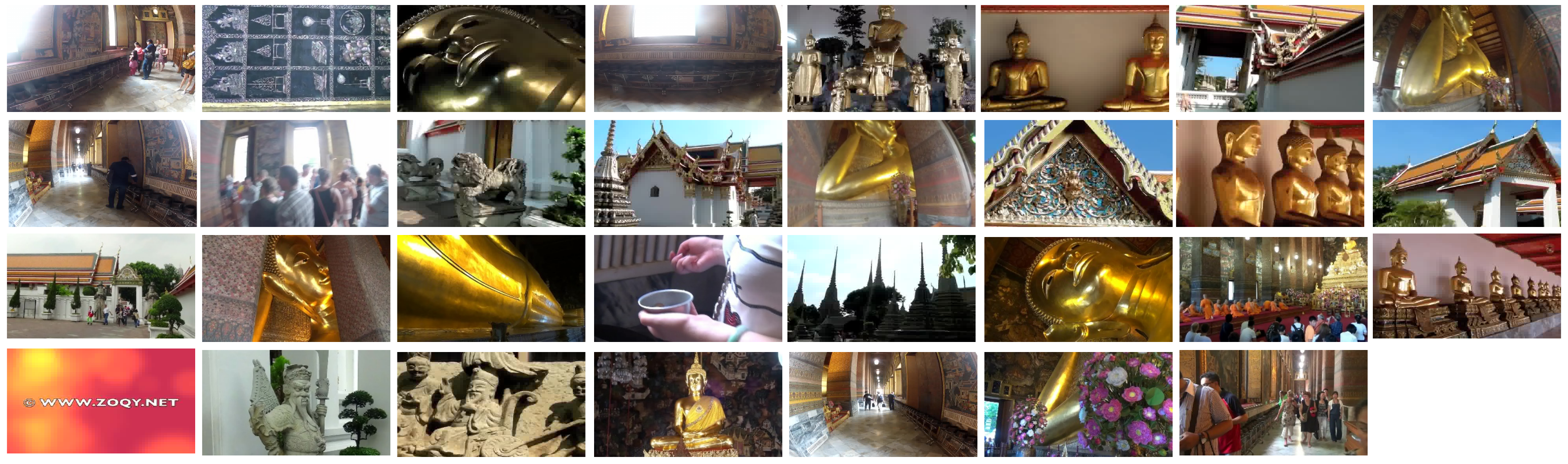}
	\end{tabular}
	\vspace{-4mm}
	\caption
	{Representative video segments generated by our approach (\textsf{DiMS}) in summarizing videos of the tourist attraction Wat Pho. We show the summaries at 10\% length and represent each summarized segment using the corresponding central frame. As can be seen, our approach generates a summary that visualizes most of the concepts related to Wat Pho. Our approach achieved the highest F-measure of 0.722 compared to 0.625 by the \textsf{MultiVideoContent} baseline.}
	\label{fig:WP} 
	\vspace{-5mm}
\end{figure*}

\begin{figure}
	\centering
	\begin{tabular}{c}
		\includegraphics[width=0.97\linewidth]{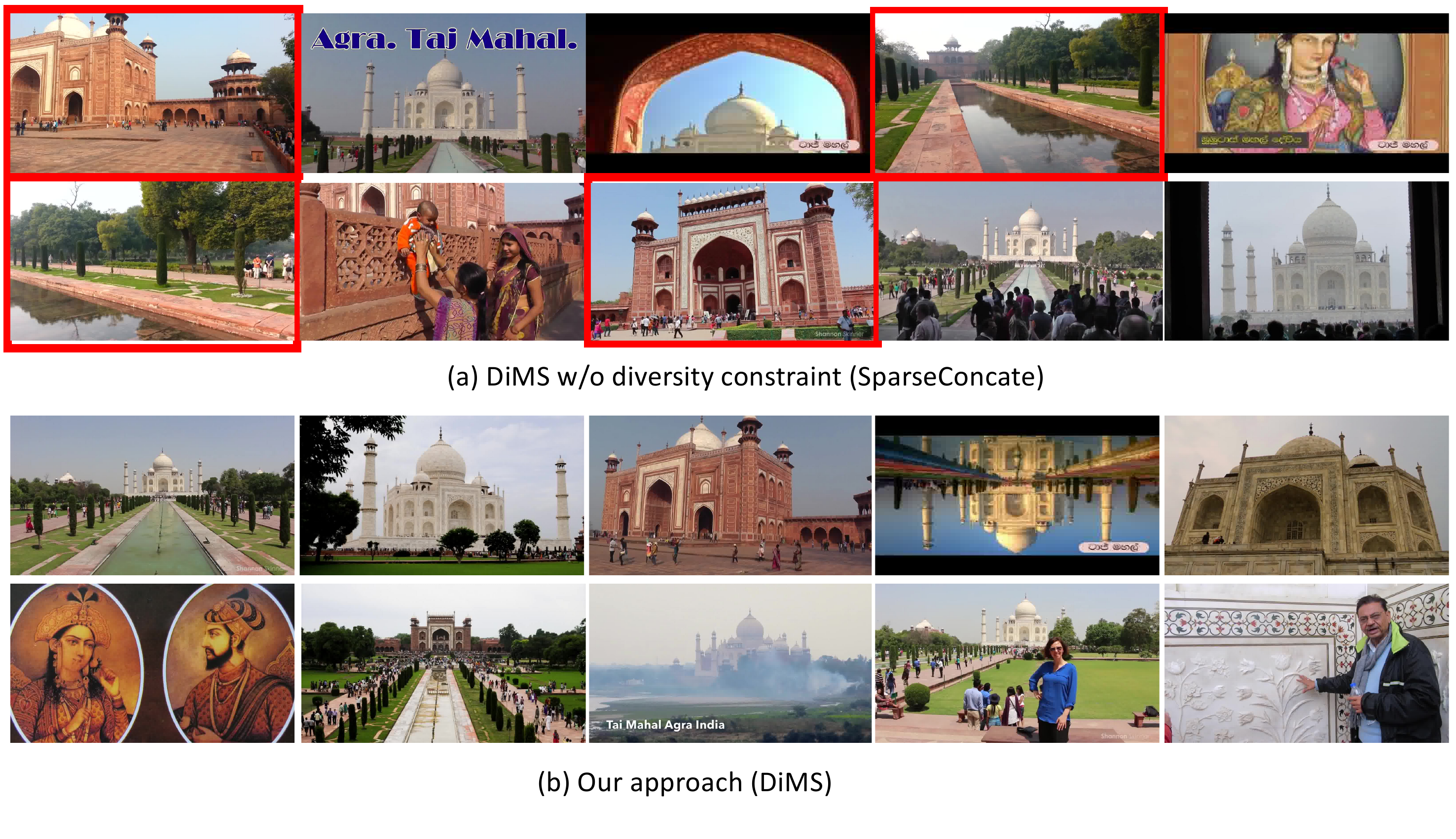}
	\end{tabular}
	\vspace{-5mm}
	\caption
	{  
	Role of diversity constraint in summarizing videos of Taj Mahal. (a) \textsf{DiMS} w/o diversity constraint (i.e., \textsf{SparseConcate} baseline), and (b) Our approach (\textsf{DiMS}). We show the top 10 segments generated using 10\% summary length. 
	As can be seen from (a), \textsf{SparseConcate} baseline finds redundant segments (marked with red color boarders) since it does not consider diversity of multiple videos. 
	Our approach \textsf{DiMS}, in contrast, generate a more informative summary capturing different but also important information described in the videos by exploring the complementary information.
	}
	\label{fig:DiversityCon1} 
	\vspace{-3mm}
\end{figure}

\vspace{1mm}
\underline{\textit{Comparision with Single-Video Baseline Methods.}}
Table~\ref{tab:CoSum1} shows the mean F-measure at 10\% summary length on Tour20 dataset. While comparing with the single-video baseline methods, we have the following key findings from Table~\ref{tab:CoSum1}: (1) The proposed method, \textsf{DiMS} statistically significantly outperforms all the compared single-video summarization methods ($p<.01$).
We observe that directly applying these methods
to summarize multiple videos produces a lot of redundant
segments which deviates from the fact that the optimal summary should
be diverse and can describe the
multi-video concepts. This is probably because these methods
are specific to single-video summarization and thus can
not take the advantage of the complementary information
among multiple videos. 
(2) Among the alternatives, the \textsf{SubMod} baseline is the most competitive. However, the gap is still significant due to the fact that the proposed optimization approach efficiently explores the complementary information in creating an optimal summary from multiple videos. The mean F-measure performance improvements over \textsf{SubMod} is about 10\% (0.613 vs 0.512) on our newly introduced Tour20 dataset.
(3)  Furthermore, note that our approach \textsf{DiMS} outperforms the na{\"i}ve approach, \textsf{SparseConcate}, that summarizes multiple videos without any constraint with a clear margin (0.613 vs 0.503). This explicitly corroborates the effectiveness of our proposed diversity regularization (Eq.~\ref{eq:diversity}) in creating an informative and compact multi-video summary (See Fig.~\ref{fig:DiversityCon1} for an illustrative example).  
(4)  Our approach outperforms both of the graph clustering based methods (\textsf{Graph}, \textsf{DT}) by a significant margin due to its ability to efficiently model multi-video correlations.


\vspace{1mm}
\underline{\textit{Comparision with State-of-the-art Methods.}} 
Table~\ref{tab:CoSum} shows the topic-wise mean F-measure performance of our method along with two multi-video summarization methods on Tour20 dataset. Following observations can be made from Table~\ref{tab:CoSum}:
(1) Our method
achieves the highest overall score of 0.613, while the strongest
baseline reaches 0.517 on the Tour20 dataset. Our approach is
able to find the important segments from a video collection which
are comparable to manual human created summaries (See Fig.~\ref{fig:WP}).
(2) Surprisingly, the performance of \textsf{SubMod} baseline is superior compared to \textsf{MultiVideoContent}. It is probably because \textsf{SubMod} considers both interestingness and representativeness in summarizing videos whereas the later one only optimizes for representativeness which may leave out some interesting segments in the summary.
(3) Our method overall produces better summaries by optimizing all the important criteria of a video summary as explained earlier. However, it has a lower performance for certain videos, e.g., videos of the topic \enquote{Burj Khalifa}. These videos contain fast motion and subtle semantics that define important segments of the video, such as opening the parachute or a nice panning shot from the top of the building. We believe these are difficult to capture without an additional semantic analysis~\cite{mei2013near}; we leave this as an interesting future work.


\begin{figure}
	\centering
	\begin{tabular}{c}
		\includegraphics[width=0.97\linewidth]{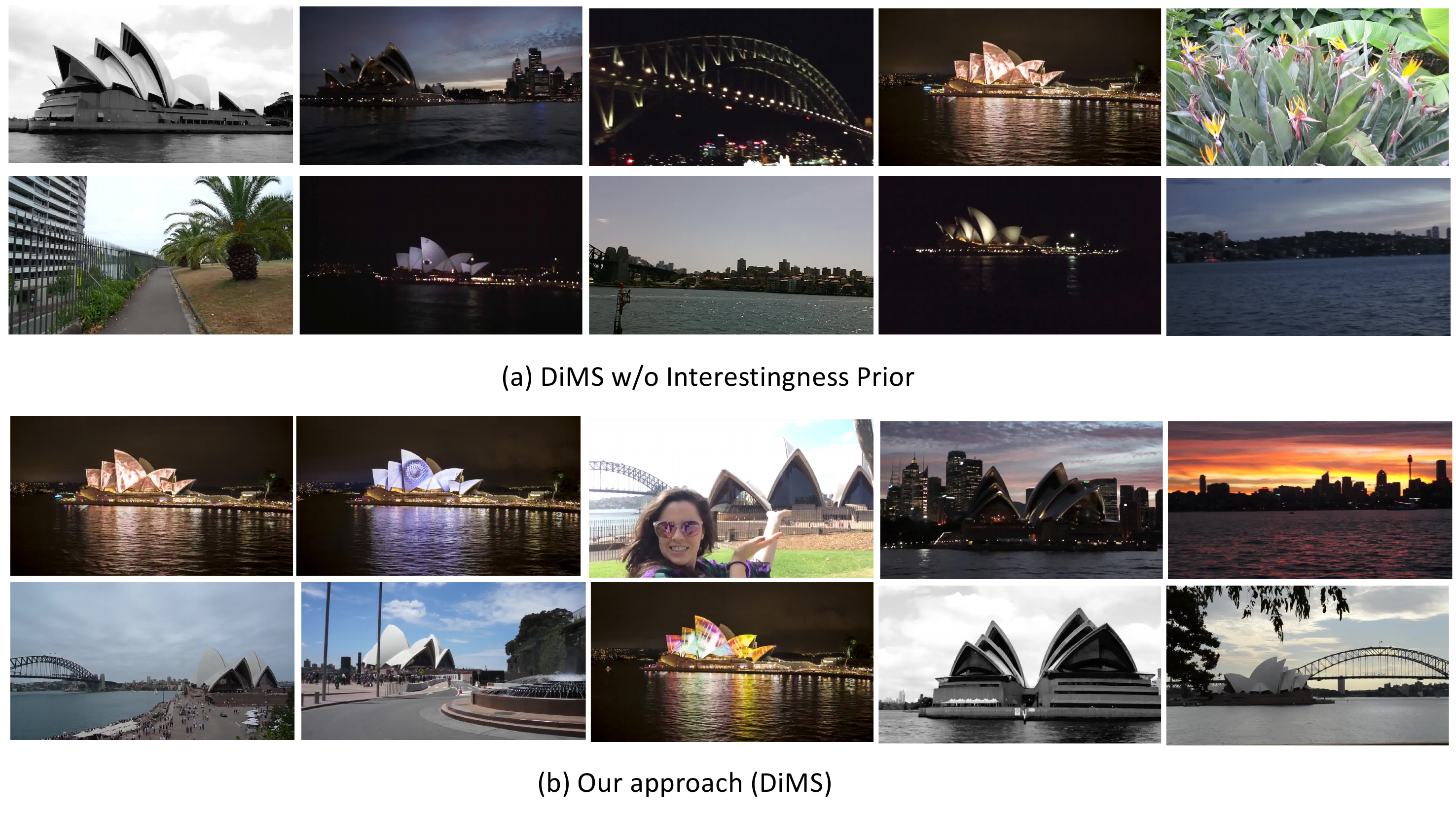}
	\end{tabular}
	\vspace{-5mm}
	\caption
	{  
		Role of interestingness prior in summarizing videos of Sydney Opera House. (a) \textsf{DiMS} w/o interestingness prior (by setting $Q^{(v)}=I$ in problem (\ref{eq:multivideo})), and (b) Our approach (\textsf{DiMS}). 
		We show the top segments generated using 10\% summary length.
		As can be seen, optimizing only for representativeness misses some crucial segments (e.g., the girl taking a photo by pointing to the opera house or segments showing several persons roaming around the house), which are indeed captured in our summary by jointly considering both representativeness and interestingness in the sparse optimization.
	}
	\label{fig:InterestAdvantage} 
	\vspace{-4mm}
\end{figure}

\vspace{1mm}
\underline{\textit{Performance Analysis with C3D Features:}} We investigate the importance and reliability of C3D features by comparing with 2D segment-level deep features, and found that the later produces inferior results, with a mean F-measure of 0.572 compared to 0.613 by the C3D features. We utilize Pycaffe with the VGG net pretrained model~\cite{simonyan2014very} to extract a 4096-dim feature vector of a frame and then use temporal mean pooling to compute a single segment-level feature vector, similar to C3D features described in~Sec. \ref{sec:video representation}. The spatio-temporal C3D features perform best, as they exploit the temporal aspects of activities typically shown in videos.

\vspace{1mm}
\underline{\textit{Performance Analysis with Interestingness Prior:}} To better understand the contribution of interestingness prior in summarizing videos, we analyzed the performance of the proposed approach by setting $Q^{(v)}=I$ in problem (\ref{eq:multivideo}), where $I$ denote the identity matrix of appropriate dimension. By turning off the interestingness prior, the mean F-measure decreases to 0.556. This is due to the fact that sparse representative selection in (\ref{eq:singlevideo}) only consider reconstruction capability and sparsity in summarizing videos. Optimizing only for representativeness risks leaving out some crucial segment(s) which are indeed captured in the summary by combining both interestingness and representativeness in summarizing videos (See Fig.~\ref{fig:InterestAdvantage} for an example). So, we conjecture that interestingness is also an important factor in summarization to generate a more condensed, descriptive and aesthetically pleasing summary.

\vspace{1mm}
\underline{\textit{Performance Analysis with Diversity Constraint:}} Fig.~\ref{fig:DiversityCon1} shows the advantage of our proposed diversity regularization in summarizing videos of Taj Mahal. By turning off the diversity constraint (i.e., \textsf{SparseConcate} baseline), the mean F-measure decreases from 0.613 to 0.503 on Tour20 dataset. Furthermore,  we also compare our approach with importance weighted clustering methods~\cite{dhillon2004kernel,modha2003feature}, i.e., \textsf{ConcateWKmeans}, \textsf{ConcateWSpectral}, \textsf{KmeansWConcate} and \textsf{SpectralWConcate} to explicitly show the advantage of our diversity constraint in generating informative summaries. We use the interestingness score of each video segment for weighting the segment-level C3D features and then perform clustering on the feature weighted space to generate video summaries~\cite{dhillon2004kernel,modha2003feature,de2012minkowski}. We use the same method~\cite{LucVanGool2014} to compute the interestingness score of each video segment--such a setting gives a fair comparison in our experiments. Table~\ref{tab:CoSumW} shows the comparison with importance weighted clustering methods on Tour20 dataset. We have the following key findings from Table~\ref{tab:CoSumW}: The performance of importance weighted clustering methods are superior compared to the classical K-means and spectral clustering (a maximum improvement of about 3\%). This is expected since the weighted version of K-means and spectral clustering use the interestingness prior while summarizing videos. However, the proposed method, \textsf{DiMS} still outperforms these methods by a significant margin which again shows the advantage of our proposed diversity regularization in selecting informative and diverse segments from a video collection.  

\begin{table} [t]
	\tiny
	\caption{Comparison with importance weighted clustering methods. Numbers show mean F-measures at 10\% summary length.}
	\label{tab:CoSumW}
	\vspace{-2mm}
	\begin{tabulary}{1.1\linewidth}{|p{8mm}||P{12.5mm}|P{12.5mm}|P{12.5mm}|P{12.5mm}|P{4mm}|}
		\hline
		\textbf{F-measure} & \textsf{ConcateWKmeans}  & \textsf{ConcateWSpectral}  & \textsf{KmeansWConcate} & \textsf{SpectralWConcate} & \textsf{DiMS} \\
		\hhline{|-|-|-|-|-|-|}
		\textbf{mean} & 0.426	& 0.448  & 0.472 & 0.483  & \textbf{0.613}\\
		\hline 		
	\end{tabulary} \vspace{-2mm}
\end{table}

\vspace{1mm}
\underline{\textit{Performance Analyis with SLH Algorithm:}} We examined the performance of our approach using cosine similarity instead of SLH algorithm in computing segment-level correlations and found that the later produces inferior results, with a mean F-measure of 0.471 compared to 0.613 with the SLH algorithm. We kept all the parameters fixed in both of the case. This is probably because SLH algorithm tries to maintain the consistency in computing inter-video correlations via the exclusion principle~\cite{Scott1991,torki2010one} which preserves the spatial arrangement of each video in computing such correlations. On the other hand, cosine similarity does not obey the exclusion principle which results in removing some crucial segments in the summary. 
However, we also believe that learning these correlations (as a future work) via a Siamese network or multiple kernel learning will further enhance our performances. 

\begin{table*} [t]
	\tiny
	\centering
	\caption{Performance comparison with several baselines including both single and multi-view methods applied on the three multi-view datasets. All the reported values are in percentage. Ours perform the best.
	} 		
	\label{tab:Multi-view}
	\begin{tabulary}{1.1\linewidth}{|p{20mm}|P{10mm}|P{10mm}|P{14mm}|P{10mm}|P{10mm}|P{14mm}|P{10mm}|P{10mm}|P{14mm}|}
		\hline
		
		& \multicolumn{3}{c|}{\textbf{Office}} &\multicolumn{3}{c|}{\textbf{Campus}}&\multicolumn{3}{c|}{\textbf{Lobby}}\\
		\cline{2-10}
		
		\textbf{Methods} &\textbf{Precision}& \textbf{Recall} & $\bf{F-measure}$ &\textbf{Precision}& \textbf{Recall} & $\bf{F-measure}$ &\textbf{Precsion}& \textbf{Recall} & $\bf{F-measure}$ \\
		\hhline{|=|=|=|=|=|=|=|=|=|=|}
		{\textsf{ConcateKmeans}}	& 100	& 38	& 55.07 & 55 & 41 & 47.06 & 85	& 67	& 75.05 \\
		{\textsf{ConcateSpectral}} & 100	& 54 & 66.99 &59 &45  &50.93  & 93	& 65 & 76.69	\\
		{\textsf{ConcateSparse}}	& 100 	& 46 & 63.01 &62 &55  & 58.61 &86 &70 &77.18 \\
		{\textsf{KmeansConcate}}	& 100	& 53  & 68.17	& 56 & 55 & 55.70 	& 91 & 70 & 78.75 \\
		{\textsf{SpectralConcate}} & 100	& 50  &  66.67	& 54 & 52 & 52.63 	& 88 &70  & 77.93\\
		{\textsf{SparseConcate}}	& 93	& 58  & 71.30 & 56	& 62 & 58.63 & 97&67  &79.45 \\
		{\textsf{Graph}} & 100	& 50  &  66.67	& 56 & 48 &  51.86	& 91 &67  & 77.33\\
		\hhline{|=|=|=|=|=|=|=|=|=|=|}
		{\textsf{RandomWalk}}	& 100	& 61 & 75.77 & 70	& 55 & 61.56 & 100	& 77 & 86.81	\\ 
		{\textsf{RoughSets}}	& 100	& 61  & 75.77 & 69 & 57 & 62.14 & 97 & 74&84.17\\
		{\textsf{BipartiteOPF}} & 100	& 69 & 81.79 & 75	& 69 & 71.82 & 100 & 79 & 88.26	\\
		\hhline{|=|=|=|=|=|=|=|=|=|=|}
		{\textsf{DiMS(ours)}}	& \textbf{100}	& \textbf{77} & \textbf{86.91} & \textbf{83}	& \textbf{69} & \textbf{75.47} & \textbf{100} & \textbf{86} & \textbf{92.52}\\ \hline 		
	\end{tabulary} 
\end{table*}

\begin{figure*}[!t]
	\centering
	{
		\includegraphics[width=1\linewidth]{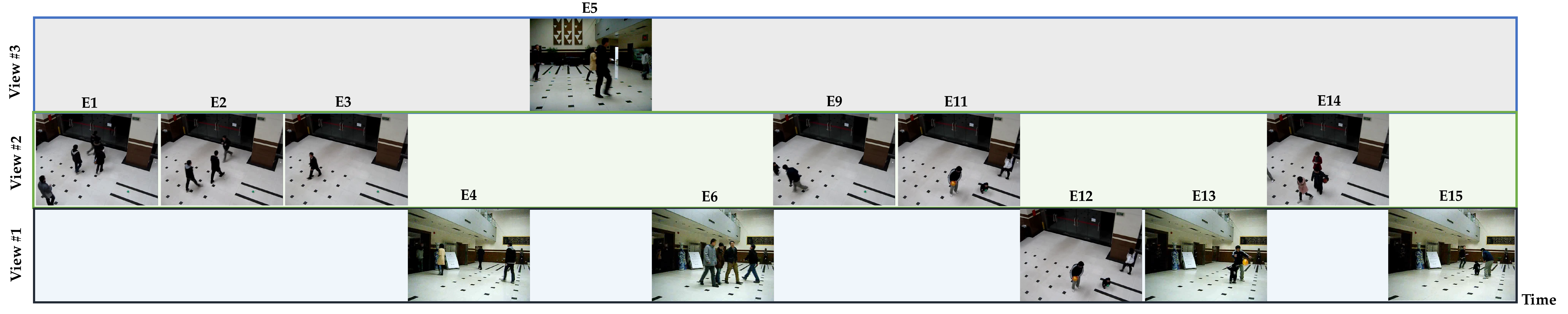}}
	\vspace{-7mm}
	\caption
	{Some summarized events for the Lobby dataset. X-axis denotes the time line as per the ground truth and the Y-axis represent the view (camera) from which the event is detected. Each event is represented by a key frame and an event number. The sequence of events in our summary are: E1: Five persons walk across the lobby towards the gate; a man runs to the gate, E2: Two men walks across the lobby towards the gate, and a man walks into the lobby, E3: A man run into the lobby from the gate, E4: Four persons walk into the lobby from the gate, E5: A man walks across the lobby towards the gate, E6: Three men are walking across the lobby towards the gate, E9: A man plays a ball with a baby, E11: A woman wearing a white coat walks across the lobby towards the gate, E12: A woman with a white coat passes away while a man is playing with a baby, E13: A man throws the ball towards the baby, E14: Two women and a man walk across the lobby from the gate, E15: A man plays a ball with a baby, a man with a black coat passes away. 
	}
	\label{fig:View-board}\vspace{-3mm}
\end{figure*} 

\vspace{-3mm}
\subsection{Multi-View Video Summarization in a Camera Network}
\underline{\textit{Goal:}} This experiment aims at evaluating our proposed framework in summarizing multi-view videos captured using a network of cameras with considerable overlapping field of views. Such a summary can be very beneficial in surveillance systems equipped in offices, banks, factories, and crossroads of cities, for obtaining significant information in short time. 

\vspace{1mm}
\underline{\textit{Datasets:}} We conduct experiments using three publicly available datasets\footnote{[Online] Available: http://cs.nju.edu.cn/ywguo/summarization.html}: (i) Office dataset captured with 4 stably-held web cameras in an indoor environment, (ii) Campus dataset taken with 4 hand-held ordinary video cameras in an outdoor scene, (iii) Lobby dataset captured with 3 cameras in a large lobby area. 

\vspace{1mm}
\underline{\textit{Performance Measures.}} We use three quantitative
measures on all experiments, including Precision, Recall
and F-measure~\cite{MultiviewTMM2010,SanjaySir2015}. For all these metrics, the higher value indicates better summarization quality. 

\vspace{1mm}
\underline{\textit{Compared Methods.}} We contrast our approach with total of ten existing approaches including seven baseline methods (\textsf{ConcateKmeans}~\cite{arthur2007k}, \textsf{ConcateSpectral}~\cite{von2007tutorial}, \textsf{ConcateSparse}~\cite{Ehsan2012}, \textsf{KmeansConcate}~\cite{arthur2007k}, \textsf{SpectralConcate}~\cite{von2007tutorial}, \textsf{SparseConcate}~\cite{Ehsan2012}, \textsf{Graph}~\cite{peng2006clip}) that use single-view summarization approach over multi-view videos to generate summary and four state-of-the-art methods (\textsf{RandomWalk}~\cite{MultiviewTMM2010}, \textsf{RoughSets}~\cite{MultiviewICIP2011}, \textsf{BipartiteOPF}~\cite{SanjaySir2015}) which are specifically designed for multi-view video summarization. Similar to the experiments in topic-oriented video summarization, the first seven single-view baselines generate a multi-view summary by either applying the method to each video separately or concatenating all the videos into a single video. 
\begin{figure} 
	\centering
	{
		\includegraphics[width=1\linewidth]{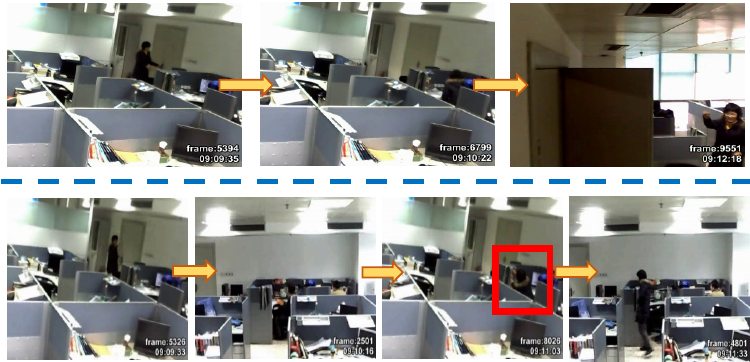}}
	\vspace{-6mm}
	\caption
	{Sequence of events detected related to activities of a member $(A_0)$ inside the Office dataset. Top row: Summary produced by method~\cite{MultiviewTMM2010}, and Bottom row: Summary produced by our approach. 
		Sequence of events detected in top row: 1st: $A_0$ enters the room, 2nd: $A_0$ sits in cubicle 1, 3rd: $A_0$ leaves the room. Sequence of events detected in bottom row: 1st: $A_0$ enters the room, 2nd: $A_0$ sits in cubicle 1, 3rd: $A_0$ is looking for a thick book to read (as per the ground truth in~\cite{MultiviewTMM2010}), and 4th: $A_0$ leaves the room. 
		The event of looking for a thick book to read (as per the ground truth in~\cite{MultiviewTMM2010}) is missing in the summary produced by method~\cite{MultiviewTMM2010} where as it is correctly detected by our approach (3rd frame: bottom row). This indicates our method captures video semantics in more informative way compared to~\cite{MultiviewTMM2010}.   
	}
	\label{fig:Event Detection} 
	\vspace{-5mm}
\end{figure} 

\vspace{1mm}
\underline{\textit{Experimental Settings.}} We set the same summary length as in~\cite{MultiviewTMM2010} to generate our summaries and then employ the ground truth of events reported in~\cite{MultiviewTMM2010} to compute the performance measures. 
We implement all the single-video summarization methods with the same video segmentation and feature representation as ours, whereas for the multi-view methods, we use prior published numbers when possible. In particular, for the multi-view summarization methods (\textsf{RandomWalk}, \textsf{BipartiteOPF}), we report the available results
from the corresponding papers and implement \textsf{RoughSets} ourselves using the same video representation as the proposed one and tune their parameters to have the best performance. 

\vspace{1mm}
\underline{\textit{Comparision with Single-View Baseline Methods:}}
Table~\ref{tab:Multi-view} shows the results on three multi-view datasets, namely Office, Campus and Lobby datasets. We have the following observations from Table~\ref{tab:Multi-view}:
(1) As expected, summaries produced using the single video-summarization methods, including the graph clustering based method (\textsf{Graph}) contain a lot of redundancies (simultaneous presence of most of the events) since they fail to exploit the complicated interview
content correlations present in multi-view videos.
(2) By using our diversity-aware sparse optimization method, such redundancy is largely reduced in contrast. Our proposed framework significantly outperforms all the single-view baseline methods in terms of precision, recall and F-measure due to its ability to model multi-view correlations.

\vspace{1mm}
\underline{\textit{Comparision with State-of-the-art Methods:}}
While comparing with state-of-the-art multi-view summarization methods, we have the following observations from Table~\ref{tab:Multi-view}:
(1) Our approach produces summaries with same precision as \textsf{RandowWalk} and \textsf{BipartiteOPF} for both Office and Lobby datasets. 
However, the improvement in recall value indicates the ability of our method in keeping more important information in the summary compared to both of the approaches (See Fig.~\ref{fig:Event Detection} for one such example). 
The performance improvements over the recently published baseline \textsf{BipartiteOPF}, on three datasets are 5.12\%, 3.65\%, 4.26\% in terms F-measure, respectively.  
(2) Notice that for all methods, including ours, performance on Campus dataset is not that good as compared to other two datasets. 
This is obvious since the Campus dataset contains many trivial events as it was captured in an outdoor environment, thus making the summarization more difficult. 
Nevertheless, for this challenging dataset, F-measure of our approach is about 4\% better than that of the recent \textsf{BipartiteOPF} and 14\% better than that of \textsf{RandomWalk}.     
Overall, on all datasets, our approach outperforms all the baselines in terms of F-measure.
This corroborates the fact that our approach produces more informative multi-view summaries in contrast to the state-of-the-art methods.        
We present a part of the summarized events for the Lobby dataset in Fig.~\ref{fig:View-board}. 

\begin{figure} [h]
	\centering
	{
		\includegraphics[width=1\linewidth]{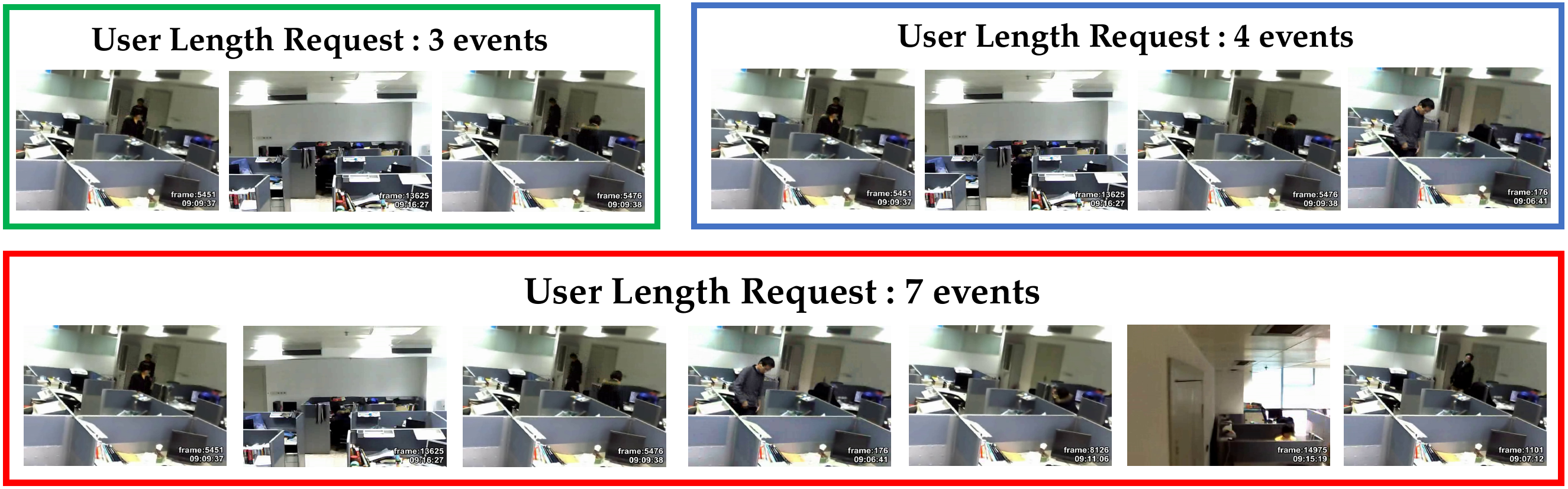}}
	\vspace{-6mm}
	\caption
	{The figure shows an illustrative example of scalability in generating summaries of different length based on the user constraints for the Office dataset. Each video segment is represented by a key frame and are arranged according to the summary generation rules mentioned in Sec. \ref{sec:Summary Generation}. 
	}             
	\label{fig:Scalability} \vspace{-2mm}
\end{figure}

\vspace{1mm}
\underline{\textit{Scalability in Generating Summaries:}}
Scalability in generating summaries of different length has shown to be effective while summarizing single videos~\cite{herranz2010framework,panda2014scalable}. However, most of the previous multi-video summarization methods~\cite{li2010multi,SanjaySir2015} require the number of representative segments to be specified before generating the summaries which is highly undesirable in practical applications. Concretely speaking, the algorithm need to be rerun for each change in the number of representative segments that the user want to see in the summary. By contrast, our approach provides scalability in generating summaries of different length based on the user constraints without any further analysis of the input videos, similar to~\cite{herranz2010framework}. This is due to the fact that a ranked list of video segments can be generated after the alternating minimization which can produce summaries of desired length without incurring any additional cost. Such a scalability property makes our approach more suitable in providing human-machine interface where the summary length is changed as per the user request. Fig.~\ref{fig:Scalability} shows the generated summaries of length 3, 4 and 7 most important events for the Office dataset.
\vspace{-2mm} 


%% file: Conclusions.tex
We present an unsupervised framework for multi-video summarization by exploring the complementarity within the videos.
We achieve this by developing a diversity-aware sparse optimization method that jointly summarizes a set of videos to find a single summary that is both interesting and representativeness of the input video collection. 
We also introduced a new dataset, Tour20, along with clear ground truth summaries to evaluate summarization algorithms in a fast and repeatable manner. 
We obtain excellent experimental results in two video summarization tasks such as topic-oriented video summarization and multi-view video summarization in a camera network, showing that our approach generates high quality summaries compared to the state-of-the-art methods. 

In our current work, we assume that videos given by a web search are relevant to the topic. However, in most practical cases, videos retrieved from search engines with topic name as a query may contain outliers and irrelevant videos due to inaccurate query text and polysemy. 
One feasible choice is to use either clustering~\cite{jing2006igroup} or additional video meta data to refine the results. Using active learning or deep CNNs~\cite{ganimproving} to get a set of topic-relevant videos is also another possibility in this regard. Moving forward, we would like to improve our method by using clustering~\cite{jing2006igroup} to handle such real-world scenarios while summarizing topic-related web videos. 
Moreover, we would like to improve our method by utilizing other types of metadata (e.g., social media images, comments, audio) while summarizing web videos.